\documentclass[10pt,twocolumn,letterpaper]{article}

\usepackage{wacv}
\usepackage{times}
\usepackage{graphicx}
\usepackage{amsmath}
\usepackage{amssymb}
\usepackage{xspace}
\usepackage{booktabs}
\usepackage{enumerate}
\usepackage{amsthm}
\usepackage{url}
\usepackage{balance}
\usepackage{algorithmic}
\usepackage{mathtools}

 \wacvfinalcopy
 
\def\wacvPaperID{127} 

\ifwacvfinal\fi

\newtheorem{theorem}{Theorem}[section]
\newtheorem{lemma}[theorem]{Lemma}
\newtheorem{proposition}[theorem]{Proposition}

\newtheorem{definition}[theorem]{Definition}

\def\Vec#1{{\boldsymbol{#1}}}
\def\Mat#1{{\boldsymbol{#1}}}

\begin{document}

\clearpage

\title{Random Projections on Manifolds of Symmetric Positive Definite Matrices\\for Image Classification}

\author
  {
  {\it Azadeh Alavi, Arnold Wiliem, Kun Zhao, Brian C. Lovell, Conrad Sanderson}\\
  ~\\
  NICTA, GPO Box 2434, Brisbane, QLD 4001, Australia\\
  University of Queensland, School of ITEE, QLD 4072, Australia\\
  Queensland University of Technology, Brisbane, QLD 4000, Australia
  }

\maketitle

\thispagestyle{empty}


\begin{abstract}
\vspace{-1ex}

\noindent
Recent advances suggest that encoding images through Symmetric Positive Definite (SPD) matrices
and then interpreting such matrices as points on Riemannian manifolds
can lead to increased classification performance.
Taking into account manifold geometry  is typically done via
(1) embedding the manifolds in tangent spaces,
or
(2) embedding into Reproducing Kernel Hilbert Spaces (RKHS).
While embedding into tangent spaces allows the use of existing Euclidean-based learning algorithms,
manifold shape is only approximated which can cause loss of discriminatory information.
The RKHS approach retains more of the manifold structure,
but may require non-trivial effort to kernelise Euclidean-based learning algorithms.
In contrast to the above approaches,
in this paper we offer a novel solution that allows SPD matrices to be used with unmodified Euclidean-based learning algorithms,
with the true manifold shape well-preserved.
Specifically, we propose to project SPD matrices using a set of random projection hyperplanes over RKHS into a random projection space,
which leads to representing each matrix as a vector of projection coefficients.
Experiments on face recognition, person re-identification and texture classification
show that the proposed approach outperforms several recent methods,
such as Tensor Sparse Coding, Histogram Plus Epitome,
Riemannian Locality Preserving Projection and Relational Divergence Classification.

\end{abstract}\vspace{-1ex}
\section{Introduction}
\label{sec:introduction}

Covariance matrices have recently been employed to describe images and videos~\cite{Pennec_jmiv06,SR_Riemannian_AVSS_2010,Tuzel_2008_PAMI},
as they are known to provide compact and informative feature description \cite{cherian2011efficient,DPM_CVPR_2011}.
Non-singular covariance matrices are naturally symmetric positive definite matrices (SPD) which form connected Riemannian manifolds when endowed with a Riemannian metric over tangent spaces~\cite{lang1999fundamentals}.
As such, the Riemannian geometry needs to be considered for solving learning tasks~\cite{Tuzel_2008_PAMI}. 

One of the most widely used metrics for SPD matrices is the Affine Invariant Riemannian Metric (AIRM)~\cite{Pennec_jmiv06}. 
The AIRM induces Riemannian structure which is invariant to inversion and similarity transforms.
Despite its properties, learning methods using this approach have to deal with computational challenges,
such as employing computationally expensive non-linear operators.

To address the above issue, two lines of research have been proposed:
(1)~embedding manifolds into tangent spaces~\cite{lui2011tangent,Porikli:2006:CT,Sanin_ICIP_2012,Tuzel_2008_PAMI,veeraraghavan2005matching};
(2) embedding into Reproducing Kernel Hilbert Spaces (RKHS), induced by kernel functions~\cite{alavi2013relational,HAMM_NIPS2009,Harandi_ICCV_2013,harandikernel,Sanin_WACV_2013,Shawe-Taylor:2004:KMP,sra2011generalized}.
The former approaches in effect map manifold points to Euclidean spaces,
thereby enabling the use of existing Euclidean-based learning algorithms.
This comes at the cost of disregarding some of the manifold structure.
The latter approach addresses this by implicitly mapping points on the manifold into RKHS, which can be considered as a high dimensional Euclidean space.
Training data can be used to define a space that preserves manifold geometry~\cite{harandikernel}.
The downside is that existing Euclidean-based learning algorithms need to be kernelised, which may not be trivial.
Furthermore, the resulting methods can still have high computational load, making them impractical to use in more complex scenarios.

\noindent
{\bf Contributions.}
In this paper we offer a novel approach for analysing SPD matrices which combines the main advantage of tangent space approaches with the discriminatory power provided by kernel space methods.
We adapt a recent idea from techniques specifically designed for learning tasks in very large image datasets~\cite{gong2012,kulis2009kernelized}.
In this domain, image representations are mapped into a reduced space wherein the similarities are still well-preserved~\cite{kulis2009kernelized}.
In our proposed approach, we employ such a mapping technique to create a space which preserves the manifold geometry while can be considered as Euclidean. 

Specifically, we first embed SPD manifold points into RKHS via the Stein Divergence Kernel~\cite{Sra_JMLR_2012}.
We then generate random projection hyperplanes in RKHS and project the embedded points via the method proposed in~\cite{kulis2009kernelized}.
Finally, as the underlying space can be thought as Euclidean, any appropriate Euclidean-based learning machinery can be applied.
In this paper, we study the efficacy of this embedding method for classification tasks.
We show that the space is only as effective as the completeness of the training data generating the random projection hyperplanes,
and address this through the use of synthetic data to augment training data.
Experiments on several vision tasks (person re-identification, face recognition and texture recognition),
show that the proposed approach outperforms several state-of-the-art methods.

We continue the paper as follows.
Section~\ref{sec:LogDetDivergance} provides a brief overview of the manifold structure and its associated kernel function.
We then detail the proposed approach in Section~\ref{sec:KLSH}.
Section~\ref{sec:Experiments_RDC} presents results on the study of the random projection space discriminability as well as comparisons with the state-of-the-art results in various visual classification tasks.
The main findings and possible future directions are summarised in Section~\ref{sec:conclusions}.
\section{Manifold Structure and Stein Divergence}
\label{sec:LogDetDivergance}

Consider $\{ \Mat{X}_1 \dots \Mat{X}_n\}\in {Sym}_{+}^{d}$ to be a set of non-singular $d{\times}d$-sized covariance matrices, which are symmetric positive definite (SPD) matrices.
These matrices belong to a smooth differentiable topological space, known as SPD manifolds.  
In this work, we endow the SPD manifold with the AIRM to induce the Riemannian structure~\cite{Pennec_jmiv06}.
As such, a point on manifold $\mathcal{M}$ can be mapped to a tangent space using:

\vspace{-2ex}
\begin{small}
\begin{equation}
    \operatorname{log}_{\Mat{X_i}} {\Mat{X_j}} = {{\Mat{X_i}}^{\frac{1}{2}} \operatorname{log}({ {\Mat{X_i}}^{-\frac{1}{2}}  \Mat{X_j}  {\Mat{X_i}}^{-\frac{1}{2}}    }) \Mat{X_i}^{\frac{1}{2}}}
    \label{eqn:AIRM_log}
\end{equation}%
\end{small}%

\vspace{-1ex}
\noindent
where $\Mat{X_i}, \Mat{X_j} \in {Sym}_+^{d}$, $\Mat{X_i}$ is the point where the tangent space is located (\ie ~tangent pole) and $\Mat{X_j}$ is the point that we would like to map into the tangent space $\mathcal{T}_{\Mat{X_i}}\mathcal{M}$; 
$\operatorname{log} ( \cdot )$ is the matrix logarithm.
The inverse function of this maps points on a particular tangent space into the manifold is:

\vspace{-2ex}
\begin{small}
\begin{equation}
    \operatorname{exp}_{\Mat{X_i}} {\Mat{y}} = {{\Mat{X_i}}^{\frac{1}{2}} \operatorname{exp}({ {\Mat{X_i}}^{-\frac{1}{2}} \Mat{y}  {\Mat{X_i}}^{-\frac{1}{2}}    }) \Mat{X_i}^{\frac{1}{2}}}
    \label{eqn:AIRM_exp}
\end{equation}%
\end{small}%

\noindent
where $\Mat{X_i} \in {Sym}_+^d$ is again the tangent pole; 
$\Mat{y} \in \mathcal{T}_{\Mat{X_i}}\mathcal{M}$ is a point in the tangent space $\mathcal{T}_{\Mat{X_i}}\mathcal{M}$;
$\operatorname{exp} ( \cdot )$ is the matrix exponential.

From the above functions, we now define the shortest distance between two points on the manifold.
The distance, here called geodesic distance, is represented as the minimum length of the curvature path that connects two points~\cite{Pennec_jmiv06}:

\vspace{-4ex}
\begin{small}
\begin{equation}
     \operatorname{d}_g^2{(\Mat{X_i},\Mat{X_j})} = \operatorname{trace} \left\{ \operatorname{log}^2 (  \Mat{X_i}^{-\frac{1}{2}} \Mat{X_j}  {\Mat{X_i}}^{-\frac{1}{2}}  )      \right\}
    \label{eqn:AIRM_dg}
\end{equation}%
\end{small}%

\vspace{-2ex}
The above mapping functions can be computationally expensive.
We can also use the recently introduced Stein divergence~\cite{Sra_JMLR_2012} to determine similarities between points on the SPD manifold.
Its symmetrised form is:

\vspace{-4ex}
\begin{small}
\begin{equation}
    J_{\phi }(\Mat{X},\Mat{Y}) \triangleq  \log \left( \det \left( \frac{\Mat{X}+\Mat{Y}}{2}\right) \right)
    - \frac{1}{2}  \log \left( \det \left( \Mat{X}\Mat{Y} \right) \right)
    \label{eqn:Stein_Div}
\end{equation}%
\end{small}%

\vspace{-2ex}
\noindent
The Stein divergence kernel can then be defined as:

\vspace{-3ex}
\begin{small}
\begin{equation}
    \operatorname{K}(\Mat{X},\Mat{Y}) = \exp\{-\sigma J_{\phi }(\Mat{X}\hspace{-0.25ex},\Mat{Y}) \} 
    \label{eqn:SDKernel}
\end{equation}%
\end{small}%

\vspace{-1ex}
\noindent
under the condition of $\sigma \in \{\frac{1}{2},\frac{2}{2},...,\frac{d-1}{2}\}$
to ensure that the kernel matrix formed by Eqn.~(\ref{eqn:SDKernel}) is positive definite~\cite{Harandi_ECCV_2012}.

\section{Random Projection on RKHS}
\label{sec:KLSH}

We aim to address classification tasks, originally formulated
on the manifold, by embedding them into a random projection space, which can be considered as Euclidean, while still honouring the manifold geometry structure.
To this end, we propose to use random projection on RKHS with the aid of the Stein divergence kernel.

Random projection is an approximation approach for estimating distances between pairs of points in a high-dimensional space~\cite{achlioptas2003database}.
In essence, the projection of a point $\Vec{u} \in \mathbb{R}^d$ can be done via a set of randomly generated hyperplanes $\{ \Vec{r}_1 \dots \Vec{r}_k \} \in \mathbb{R}^d$:

\noindent
\begin{equation}
	\operatorname{f}(\Vec{u}) = \Vec{u}^{\top} \Mat{R}
\end{equation}

\noindent
where $\Mat{R} \in \mathbb{R}^{d \times k}$ is the matrix wherein each column contains a single hyperplane $\Vec{r}_i$;
$\operatorname{f}( \cdot )$ is the mapping function which maps any point in $\mathbb{R}^d$ into a random projection space space $\mathbb{R}^k$.
According to the Johnson-Lindenstrauss lemma~\cite{achlioptas2003database}, it is possible to map a set of high-dimensional points into much lower dimensional space wherein the pairwise distance between two points are well-preserved:

\begin{lemma}
	Johnson-Lindenstrauss Lemma. For any $\epsilon$ such that \mbox{$\frac{1}{2} > \epsilon > 0$}, and any set of points $S \in \mathbb{R}^d$ with $|S| = n$ upon projection to a uniform random k-dimension subspace where $k = \operatorname{O}(\operatorname{log}\ n)$, the following property holds with probability at least $\frac{1}{2}$ for every pair $\Vec{u}, \Vec{v} \in S$,
	$(1-\epsilon) || \Vec{u} - \Vec{v}||^2 \leq ||\operatorname{f}(\Vec{u}) - \operatorname{f}(\Vec{v})||^2 \leq (1+\epsilon) ||\Vec{u} - \Vec{v}||^2$, where $\operatorname{f}(\Vec{u}), \operatorname{f}(\Vec{v})$ are projection of $\Vec{u},\Vec{v}$.
\end{lemma}

Despite the success of numerous approaches using this lemma to accomplish various computer vision tasks, most of them restrict the distance function to the $\ell_p$ norm, Mahalanobis metric or inner product~\cite{charikar2002similarity,datar2004locality,jain2008fast}, which makes them incompatible for non-Euclidean spaces. 
Recently, Kulis and Grauman~\cite{kulis2009kernelized} proposed a method that allows the distance function to be evaluated over RKHS.
Thus, it is possible to apply the lemma for any arbitrary kernel $\mathbb{K}(i,j)=\operatorname{K}(\Vec{X}_i,\Vec{X}_j)=\operatorname{\phi}(\Vec{X}_i)^{\top}\operatorname{\phi}(\Vec{X}_j)$ for an unknown embedding $\operatorname{\phi}( \cdot )$ which maps the points to a Hilbert space $\mathcal{H}$~\cite{kulis2009kernelized}. 
This approach makes it possible for one to construct a random projection space on an SPD manifold, where the manifold structure is well-preserved.

The main idea of our proposed approach, denoted as {\bf R}andom Projection {\bf O}n {\bf S}PD manifold for Imag{\bf E} Classification (ROSE),
is to first map all points on the manifold into RKHS, with implicit mapping function $\operatorname{\phi}( \cdot )$, via the Stein divergence kernel.
This is followed by mapping all the points in the RKHS $\operatorname{\phi}(\Vec{\Mat{X_i}}) \in \mathcal{H}$  into a random projection space $\mathbb{R}^k$.
To achieve this we follow the Kulis-Grauman approach~\cite{kulis2009kernelized} by randomly generating a set of hyperplanes over the RKHS $\{\Vec{r}_1 \dots \Vec{r}_k\} \in \mathcal{H}$ which is restricted to be approximately Gaussian.
As the embedding function $\operatorname{\phi}( \cdot )$ is unknown, then the generation process is done indirectly via a weighted sum of the subset of the given training sets.

To this end, consider each data point $\operatorname{\phi}(\Mat{X}_i)$ from the training set as a vector from some underlying distribution $D$ with unknown mean $\Vec{\mu}$ and unknown covariance $\Mat{\Sigma}$.
Let $S$ be a set of $t$ training exemplars chosen i.i.d.~from $D$, then $\Vec{z}_t = \frac{1}{t}\sum_{i\in S} \operatorname{\phi}(\Mat{X}_i)$ is defined over $S$. 
According to the central limit
theorem for sufficiently large $t$, the random vector $\tilde{\Vec{z}_t} = \sqrt t (\Vec{z}_t - \Vec{\mu})$ is distributed according to the multi-variate Gaussian $\mathcal{N}(\Vec{\mu},\Mat{\Sigma})$~\cite{rice2007mathematical}.
Then if a whitening transform is applied, it results in $\Vec{r}_i = \Mat{\Sigma} ^{-\frac{1}{2}} \tilde{\Vec{z}}_t$ which follows $\mathcal{N}(0,\Mat{I})$ distribution in Hilbert space $\mathcal{H}$.
Therefore the $i$-th coefficient of each vector in the random projection space is defined as:

\vspace{-2ex}
\begin{equation}
	\operatorname{\phi}(\Vec{X}_i)^{T} \Mat{\Sigma}^{-\frac{1}{2}} \tilde{\Vec{z}}_t
	\label{eqn:KLSH_embedding}
\end{equation}%
\vspace{-2ex}

The mean $\Vec{\mu}$ and covariance $\Mat{\Sigma}$ need to be approximated from training data. 
A set of $p$ objects is chosen to form the first $p$ items of a reference set: $\operatorname{\phi}(\Vec{X}_1), \dots, \operatorname{\phi}(\Vec{X}_p)$.
Then the mean is implicitly estimated as $\Vec{\mu} = \frac{1}{p}\sum_{i=1}^{p} \operatorname{\phi}(\Vec{X}_i)$, and the covariance matrix $\Mat{\Sigma}$ is also formed over the $p$ samples.
Eqn.~(\ref{eqn:KLSH_embedding}) can be solved using a similar approach as for Kernel PCA, which requires projecting onto
the eigenvectors of the covariance matrix~\cite{kulis2009kernelized}. Let the eigendecomposition of $\Mat{\Sigma}$ be $UV{U}^{T}$, then $\Mat{\Sigma}^{-\frac{1}{2}} = U{V}^{\frac{1}{2}}{U}^{T}$,
and therefore Eqn.~(\ref{eqn:KLSH_embedding}) can be rewritten as~\cite{kulis2009kernelized}:

\vspace{-2ex}
\begin{equation}
	\operatorname{\phi}(\Vec{X}_i)^{T} U{V}^{\frac{1}{2}}{U}^{T} \tilde{\Vec{z}}_t
	\label{eqn:KLSH_embedding_2}
\end{equation}%
\vspace{-2ex}

Let then define $\Mat{K}$ as a kernel matrix over the $p$ randomly selected training points, where $\Mat{K} = \Lambda \Theta {\Lambda}^{T}$.
Based on the fact that the non zero eigenvalues of $V$ are equal to the non-zero eigenvalues of $\Theta$,
Kulis-Grauman~\cite{kulis2009kernelized} showed that Eqn.~(\ref{eqn:KLSH_embedding_2}) is equivalent to:

\noindent
\vspace{-1ex}
\begin{equation}
	 \sum\nolimits_{i=1}^{p} \omega(i) ( {\operatorname{\phi}(\Vec{X}_i)^{T}} {\operatorname{\phi}(\Vec{X})}) 
	\label{eqn:KLSH_embedding_3}
\end{equation}%

\vspace{-1ex}
\noindent
where 

\noindent
\vspace{-2ex}
\begin{equation}
	 \omega(i) = \frac{1}{t} \sum_{j=1}^{p} \sum_{l\in S}^{}  {{K_{ij}}}^{-\frac{1}{2}}  {{K_{jl}}} - \frac{1}{p} \sum_{j=1}^{p} \sum_{k=1}^{p}  {{K_{ij}}}^{-\frac{1}{2}}  {{K_{jk}}}
	\label{eqn:KLSH_embedding_4}
\end{equation}%

\vspace{-1ex}
\noindent
where, for $S$, a set of $t$ points are randomly selected from the $p$ sampled points. 
The expression $w(i)$ in Eqn.~(\ref{eqn:KLSH_embedding_4}) can be further simplified by defining $e$ as a vector of all ones, and $e_S$ as a zero vector with ones in the entries corresponding to the indices of $S$~\cite{kulis2009kernelized}:

\vspace{-2ex}
\begin{equation}
	 w =  {\Mat{K}}^{\frac{1}{2}} \left ( \frac{1}{t} e_s - \frac{1}{p} e \right )
	\label{eqn:KLSH_embedding_5}
\end{equation}%
\vspace{-2ex}
 
In terms of calculating the computational complexity of the training algorithm, according to Eqns.~(\ref{eqn:KLSH_embedding_3}) and~(\ref{eqn:KLSH_embedding_5}),
the most expensive step is in the single offline computation of ${\Mat{K}}^{\frac{1}{2}}$, which takes $O(p^3)$.
The computational complexity of classifying a query point depends then on three factors:
computing the kernel vector which requires $O(p d^3)$ operations, projecting the resulting kernel vector into random hyperplane which demands $O(pt)$ operations (where $t < p$),
and finally applying a classifier in the projection space which can be done with one versus all support vector machine $O(nb)$ operations,
where $n$ is the number of classes and $b$ is the number of hyperplanes used in defining the hyperplane~\cite{kulis2009kernelized}.
Hence the complexity of classification for a single query data is equal to $O(p^3+pt+nb)$ which is more efficient than Relational Divergence based Classiﬁcation (RDC)~\cite{alavi2013relational}, which is later shown to be the second best approach in the experiment part. The RDC represents Riemannian points as similarity vectors to a set of training points.
As similarity vectors are in Euclidean space, RDC then employs Linear Discriminant Analysis as a classifier.

\subsection{Synthetic Data}

As later shown in the experiment section (for instance the result shown in Fig.~\ref{fig:ETHZ_SyntheticHighlight}),
the discriminative power of the random projection space depends heavily on the training set which generates the random projection hyperplanes.
To overcome this limitation, we propose to use generated synthetic SPD matrices \mbox{\small $\Mat{X_1},...,\Mat{X_n} \in {Sym}_{+}^d$} centred around the mean of the data (denoted by $\Mat{X_{\mu}}$),
where the mean of the training set can be determined intrinsically via the Karcher mean algorithm~\cite{Pennec_jmiv06}.

We relate the synthetic data to the training set, by enforcing the following condition on the synthetic SPD matrices:

\vspace{-2ex}
\begin{small}
\begin{equation}
\forall \Mat{X_j} \in S~~and~~\forall \Mat{X_i} \in G~~:~~ \\
\label{eqn:KarcherMeanCond}
\end{equation}%
\end{small}%
\vspace{-2ex}
\begin{small}
\begin{equation*}
\operatorname{d_g}(\Mat{X_{\mu}} , \Mat{X_j}) \leqslant \operatorname{max}(\operatorname{d_g}(\Mat{X_{\mu}} , \Mat{X_i})) \nonumber
\end{equation*}%
\end{small}%
\vspace{-2ex}

\noindent
where $G$ is the training set, $S$ is a set of $t$ training exemplars chosen i.i.d. from some underlying distribution $D$, $\Mat{X_{\mu}}$ is the mean of the training set and $\Mat{X_j}$ is a generated synthetic point.

The constraint in Eqn.~(\ref{eqn:KarcherMeanCond}) considers a ball around the mean of the training data, with the radius equal to the longest calculated distance between mean and the given training points: 
\vspace{-2ex}

\begin{equation}
r = \operatorname{max}(\operatorname{d_g}(\Mat{X_{\mu}} , \Mat{X_i})) 
\label{eqn:KarcherMeanCond_r}
\end{equation}
\vspace{-2ex}

Then we need to generate SPD matrices which are located within $r$ radius from the mean  (Eqn.~\ref{eqn:KarcherMeanCond_r}). It is not trivial to generate SPD matrices which follow Eqn.~(\ref{eqn:KarcherMeanCond}),
as it establishes a relation between generate SPD matrices and the original training points.
To address this, we apply the relationship between the geodesic distance and the given Riemannian metric in a tangent space.
Let $\Mat{X_i}, \Mat{X_j} \in {Sym}_{+}^{d}$ be two points on the manifold and $\Mat{x_i}, \Mat{x_j} \in \mathcal{T}_{\Mat{X_i}}\mathcal{M}$ be the corresponding points on the tangent space $\mathcal{T}_{\Mat{X_i}}\mathcal{M}$.
The norm of vector $\overline{\Mat{x_i}\Mat{x_j}}$ is equivalent to $\operatorname{d_g}(\Mat{X_i}, \Mat{X_j})$~\cite{Pennec_jmiv06}.
Therefore, it is possible to find a point $\Mat{Y_i}$ along the geodesic $\Mat{X_i}$ and $\Mat{X_j}$ whose geodesic distance to $\Mat{X_i}$ satisfies (\ref{eqn:KarcherMeanCond}).

Along with the above definitions, we introduce the following definition and proposition:

\begin{definition}
	Any point on an SPD manifold $\Mat{X_i} \in {Sym}_{+}^{d}$ is said to have normalised geodesic distance with respect to $\Mat{X_j} \in {Sym}_{+}^{d}$ if and only if $\operatorname{d_g}(\Mat{X_i},\Mat{X_j}) = 1$.

\end{definition}

\begin{proposition}
\label{prop:1}
	For any two SPD matrices $\Mat{X}, \Mat{X_{\mu}} \in {Sym}_{+}^{d}$,
	there exists ${\Mat{X_g}}$  on the geodesic curve defined on ${\Mat{X}}$ and $\Mat{X_{\mu}}$,
	which has normalised geodesic distance with respect to $\Mat{X_{\mu}}$.
	The point ${\Mat{X_g}}$ can be determined via:
	$\Mat{X_{\mu}}^{\frac{1}{2}} \left( \Mat{X_{\mu}}^{-\frac{1}{2}} \Mat{X} \Mat{X_{\mu}}^{-\frac{1}{2}}   \right)^{c} \Mat{X_{\mu}}^{\frac{1}{2}}$,
	where $c = \frac{\zeta }{\operatorname{d_g}(\Mat{X}, \Mat{X_{\mu}})}$, for $\zeta = 1$. 
\end{proposition} 

\noindent
To prove the above proposition, we let $\Mat{X}, \Mat{X_{\mu}} \in {Sym}_{+}^{d}$ to be two given points on an SPD manifold.
In order to normalise the geodesic distance of $\Mat{X}$ with respect to $\Mat{X_{\mu}}$, we map point $\Mat{X}$ into tangent space $\mathcal{T}_{\Mat{X_\mu}}\mathcal{M}$.
As a tangent space is considered as Euclidean space where the distance between $\Mat{X}$ and tangent pole $\Mat{X_{\mu}}$ is preserved, Euclidean vector normalisation can be applied.
Finally the normalised point is mapped back to the manifold.
These steps can be presented as:

\vspace{-2ex}
\begin{small}
\begin{equation}
  {\Mat{X_g}} = \operatorname{exp}_{\Mat{X_{\mu}}} \left( \frac{\zeta}{\operatorname{d_g}(\Mat{X_{\mu}}, \Mat{X})} \operatorname{log}_{\Mat{X_{\mu}}} ( \Mat{X} ) \right)
\end{equation}%
\end{small}%

\noindent
By plugging in (\ref{eqn:AIRM_log}) and~(\ref{eqn:AIRM_exp}) we obtain:
\begin{small}
\begin{equation*}
  {\Mat{X_g}} = {\Mat{X_{\mu}}}^{\frac{1}{2}} \operatorname{exp} \left( \frac{\zeta}{\operatorname{d_g}(\Mat{X_{\mu}}, \Mat{X})} \operatorname{log} ( \Mat{X_{\mu}}^{-\frac{1}{2}} \Mat{X} \Mat{X_{\mu}}^{-\frac{1}{2}}) \right) \Mat{X_{\mu}}^{\frac{1}{2}} 
\end{equation*}%
\end{small}%

\noindent
If we let $c = \frac{\zeta}{\operatorname{d_g}(\Mat{X_{\mu}}, \Mat{X})}$, based on the fact that $\Mat{X}$ and $\Mat{X_{\mu}}$ are SPD matrices, we arrive at:%
\footnote{See the appendix for proof of $\operatorname{log}{\Mat{X}^c} \mbox{=} c~\operatorname{log}{\Mat{X}}$, where $\Mat{X} \in {Sym}_{+}^{d}$.}
\begin{small}
\begin{equation*}
  {\Mat{X_g}} = \Mat{X_{\mu}}^{\frac{1}{2}} \operatorname{exp} \left(  \operatorname{log} \left( ( \Mat{X_{\mu}}^{-\frac{1}{2}} \Mat{X} \Mat{X_{\mu}}^{-\frac{1}{2}})^c \right) \right)\Mat{X_{\mu}}^{\frac{1}{2}} 
\end{equation*}
\end{small}
\noindent
which proves that:
\begin{small}
\begin{equation}
  {\Mat{X_g}} = \Mat{X_{\mu}}^{\frac{1}{2}} \left( \Mat{X_{\mu}}^{-\frac{1}{2}} \Mat{X} \Mat{X_{\mu}}^{-\frac{1}{2}}   \right)^{c} \Mat{X_{\mu}}^{\frac{1}{2}}
\end{equation}%
\end{small}%

\noindent
Having $\zeta$ is equal to $1$ results a normalised geodesic distance with respect to $\Mat{X_{\mu}}$.
However in our case to satisfy Eqn.~(\ref{eqn:KarcherMeanCond}),
we use $\zeta = \delta \times \operatorname{max}(\operatorname{d_g}(\Mat{X_{\mu}} , \Mat{X}_j))$,
where $\delta \in [0, 1]$ is randomly generated number according to uniform distribution.

\section{Experiments and Discussion}
\label{sec:Experiments_RDC}

We consider three computer vision classification tasks:
(1) texture classification~\cite{randen1999filtering}; (2) face recognition~\cite{phillips2000feret} and (3) person re-identification~\cite{ess2007depth}.
We first detail the experiment set up for each application and discuss our results for the comprehensive study of the random projection space discriminability on the tasks.
To this end, we first embed the SPD matrices into RKHS via the Stein divergence kernel, followed by mapping the embedded data points into a random projection space.
The resulting vectors are then fed to a linear Support Vector Machine classifier, which uses a one-versus-all configuration for multi-class classification~\cite{REF08a,Shawe-Taylor:2004:KMP}

The parameter settings are as follows.
As suggested in~\cite{kulis2009kernelized}, we have used $t = \min(30,\frac{1}{4}n)$, where $n$ is the number of samples chosen to create each hyperplane.
For the number of the random hyperplanes we have used validation data to choose one of $n$, $2n$ or $3n$.
Based on empirical observations on validation sets, the number of synthetic samples was chosen as either $n$ or $m$, where $m$ be the number of samples per class.
In a similar manner, the value of $\sigma$ in Eqn.~(\ref{eqn:SDKernel}) was chosen from $\{1, 2, \dots, 20\}$.

We compare our proposed method, here denoted as {\bf R}andom Projection {\bf O}n {\bf S}PD manifold for Imag{\bf E} Classification (ROSE),
with several other embedding approaches (tangent spaces, RKHS and hashing)
as well as several state-of-the-art methods.
We also evaluate the effect of augmenting the training data with synthetic data points,
and refer to this approach as ROSE with {\bf S}ynthetic data (ROSES).

\begin{figure}[!t]
  \centering
  \includegraphics[width=0.8\columnwidth,keepaspectratio]{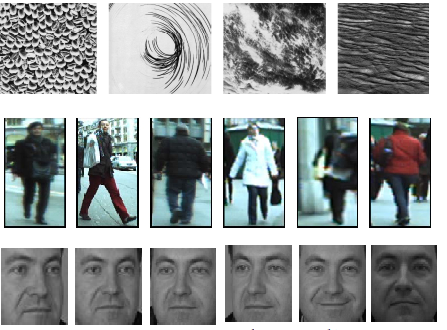}
  \centering
  \caption
    {
    \small
    Top row: example of pedestrians in the ETHZ dataset~\cite{Schwartz_ETHZ_short}.
    Middle row: example images from the Brodatz texture dataset~\cite{randen1999filtering}.
    Bottom row: examples of closely-cropped faces from the FERET `b' subset \cite{phillips2000feret}.
    }
  \label{fig:Datasets_examples}
\end{figure}

For the person re-identification task we used the modified version~\cite{Schwartz_ETHZ_short} of the ETHZ dataset~\cite{ess2007depth}.
The dataset was captured from a moving camera, with the images of pedestrians containing occlusions and wide variations in appearance.
Sequence~1 contains 83 pedestrians, and Sequence~2 contains 35 (Fig.~\ref{fig:Datasets_examples}).
Following~\cite{alavi2013relational}, we first downsampled all the images and then created the training set using 10 randomly selected images,
while the rest were used to shape the test set.
The random selection of the training and testing data was repeated 20 times.
Each image was represented as a covariance matrix of feature vectors obtained at each pixel location:

\vspace{-2ex}
\begin{footnotesize}
\begin{equation*}
F_{x,y}
\mbox{=}
\left[
  x, y, 
  R_{x,y},   G_{x,y},   B_{x,y}, 
  R_{x,y}',  G_{x,y}',  B_{x,y}', 
  R_{x,y}'', G_{x,y}'', B_{x,y}''
\right]
\end{equation*}
\end{footnotesize}

\vspace{-2ex}
\noindent
where {\small $x$} and {\small $y$} represent the position of a pixel,~while
{\footnotesize $R_{x,y}$}, {\footnotesize $G_{x,y}$} and {\footnotesize $B_{x,y}$}
represent the corresponding colour information.
$C_{x,y}'$, $C_{x,y}''$ represent the gradient and Laplacian for colour $C$, respectively.

For the task of texture classification, we use the Brodatz dataset~\cite{randen1999filtering}.
See Fig.~\ref{fig:Datasets_examples} for examples.
We follow the test protocol presented in~\cite{sivalingam2010tensor}.
Accordingly, nine test scenarios with various number of classes were generated, 
To create SPD matrices, we follow~\cite{alavi2013relational} by downsampling each image and then splitting it into 64 regions. 
A feature vector for each pixel {\small $I\left(x,y\right)$} is calculated as
{\small
$
 F(x, y)
  \mbox{~=~}
  \left[
    I\left(x,y\right),
    \left| \frac{\partial I}  {\partial x}  \right|,  \left|\frac{\partial I}  {\partial y}  \right|,
    \left| \frac{\partial^2 I}{\partial x^2}\right|,  \left|\frac{\partial^2 I}{\partial y^2}\right|
  \right] \nonumber
$}.
Each region is described by a covariance matrix formed from these vectors.
For each test scenario, we randomly select 25 covariance matrices per class to construct training set and the rest is used to create the testing set.
The random selection was repeated 10 times and the mean results are reported.

For face recognition task, the `b' subset of the FERET dataset~\cite{phillips2000feret} is used.
Each image is first closely cropped to include only the face and then downsampled (Fig.~\ref{fig:Datasets_examples}). The tests with various pose angles were created to evaluate the performance of the method.
The training set consists of frontal images with illumination, expression and small pose variations.
Non-frontal images are used to create the test sets.
Each face image is represented by a covariance matrix, where for every pixel {\small $I(x,y)$} the following feature vector is computed:

\vspace{-2ex}
\begin{small}
\begin{equation}
F_{x,y}
\mbox{=}
\left[~ I(x,y),~ x,~ y,~ |G_{0,0}(x,y)|,~ \cdots\hspace{-0.4ex},~ |G_{4,7}(x,y)| ~\right]
\end{equation}%
\end{small}%

\noindent
where {\small $G_{u,v}{(x,y)}$} is the response of a 2D Gabor wavelet centred at {\small $x,y$}.

\subsection{Random Projection Space Discriminability}
\label{subsec:Experiment_ROSE}

We first compare the performance of the proposed ROSE method with several other embedding methods:
(1)~Kernel SVM (KSVM) using the Stein divergence kernel,
(2)~Kernelised Locality-Sensitive Hashing (KLSH)~\cite{kulis2009kernelized},
and
(3)~Riemannian Spectral Hashing (RSH), a hashing method specifically designed for smooth manifolds~\cite{chaudhry2010fast}. 

\begin{table}[!b]
  \centering
  \caption
    {
    \small
    Recognition accuracy (in \%) for the person re-identification task on Seq.~1 and Seq~.2 of the ETHZ dataset; KSVM: Kernel SVM; KLSH: Kernelised Locality-Sensitive Hashing; RSH: Riemannian Spectral Hashing.
    ROSE is the proposed method, and ROSES is ROSE augmented with synthetic data.
    }
    \label{tab:ETHZ_ROSE}
    \vspace{0.5ex}    
 \small
    \begin{tabular}{lccccc}
    \toprule
    &~{\bf KSVM}~
	 &{\bf KLSH}~
    &{\bf RSH}~
    &{\bf ROSE}~
    &{\bf ROSES}~\\
    \toprule
    {\bf Seq.1}         &$72.0$        &$81.0$          &$58.5$     &${90.7}$ &${\bf 92.5}$\\
    {\bf Seq.2}         &$79.0$        &$84.0$          &$62.7$     &${91.5}$ &${\bf 94.0}$\\
    \bottomrule
    \toprule
	\textbf{Average}   &$75.5$  &$82.5$ &$60.6$ &$91.2$ &$\bf 93.2$ \\
	\bottomrule
   \end{tabular}
\end{table}
\begin{table}[!b]
  \centering
   \caption
    {
    \small
    Recognition accuracy (in \%) for the texture recognition task on BRODATZ dataset.
    }
    \label{tab:BRODATZ_ROSE}
    \vspace{0.5ex}    
\small
    \begin{tabular}{lccccc}
    \toprule
    &~{\bf KSVM}~
	 &{\bf KLSH}~
    &{\bf RSH}~
    &{\bf ROSE}~
    &{\bf ROSES}~\\
    \toprule
    {\bf 5c}         	&$99.3$	&$88.7$	&$96.6$	& $99.3$	& $\bf99.8$\\
    {\bf 5m}         	&$85.8$	&$43.6$	&$81.9$	&$\bf90.1$	&$88.4$\\
    {\bf 5v}         	&$86.2$	&$82.6$	&$76.9$	&$\bf91.6$	&$88.6$\\
    {\bf 5v2}         	&$89.4$	&$52.0$	&$80.9$	&$90.5$	& $\bf92.7$\\
    {\bf 5v3}         	&$87.4$	&$73.0$	&$79.1$	&$88.6$	&$\bf91.3$\\
    {\bf 10}         	&$81.3$	&$47.0$	&$72.5$	&$86.7$	&$\bf87.0$\\
    {\bf 10v}         	&$81.5$	&$48.0$	&$69.3$	&$88.1$	&$\bf88.5$\\
    {\bf 16c}         	&$79.6$	&$33.7$	&$65.7$	&$84.1$	&$\bf85.7$\\
    {\bf 16v}         	&$73.4$	&$35.5$	&$59.0$	&$77.1$	&$\bf79.8$\\
    \bottomrule
    \toprule
{\bf Average}			&$84.88$	&$56.0$	&$75.8$	&$88.5$	&$\bf89.1$\\
    \bottomrule
   \end{tabular}
\end{table}

\begin{table}[!b]
  \centering
  \caption
    {
    \small
    Recognition accuracy (in \%) for the face recognition task on the `b' subset of the FERET dataset.
    }
    \label{tab:FERET_ROSE}
    \vspace{0.5ex}    
 \small
    \begin{tabular}{lccccc}
    \toprule
    &~{\bf KSVM}~
	 &{\bf KLSH}~
    &{\bf RSH}~
    &{\bf ROSE}~
    &{\bf ROSES}~\\
    \toprule
    {\bf bd}         &$39.0$        &$70.0$          &$ 13.5$        &${\bf 70.5}$     &${52.0}$\\
    {\bf bg}         &$58.5$        &$\bf 80.5$          &$ 31.5$     &${\bf80.5}$ &${61.5}$\\
    \bottomrule
        \toprule
       {\bf Average}	&$48.8$	&$75.2$	&$22.5$	&$\bf 75.5$	&$56.8$\\
    \bottomrule
   \end{tabular}
\end{table}
Tables~\ref{tab:ETHZ_ROSE}, ~\ref{tab:BRODATZ_ROSE} and~\ref{tab:FERET_ROSE} report the results for each dataset.
ROSE considerably outperforms the other embedding methods on the texture and person re-identification applications,
while being on par with KLSH on the face recognition task.
This suggests that the random projection space constructed by the random hyperplanes over RKHS offers sufficient discrimination for the classification tasks.
In fact, as we use linear SVM for the classifier, the results presented here follow the theoretical results from~\cite{shi2012margin}
which suggest that the margin for the SVM classifier is still well-preserved after the random projection.

We apply the ROSES method (ROSE augmented with synthetic data) on the three tasks
in order to take a closer look at the contribution of the training data generating the random projection hyperplanes for space discriminability.
As shown in the results, there is notable improvement over ROSE in the ETHZ person re-identification as well as Brodatz texture classification datasets.
However, using synthetic points gives adverse effect on the FERET face recognition dataset.

The results suggest that the training data contributes to space discriminability.
This is probably due to the fact that each random projection hyperplane is represented as a linear combination of randomly selected training points.
As such, variations and completeness of the training data may have significant contributions to the resulting space.
The performance loss suffered on the FERET face recognition dataset is probably caused by the skewed data distribution of this particular dataset.
Hence adding synthetic points would significantly alter the original data distribution which in turn affects space discriminability.
From our empirical observation (while working with RSH), we found that all data points are grouped together when an intrinsic clustering method was applied to the the dataset.
The very poor performance of RSH on this dataset supports our view.

\begin{figure}[!t]
  \centering
  \includegraphics[width=1.01\columnwidth,keepaspectratio]{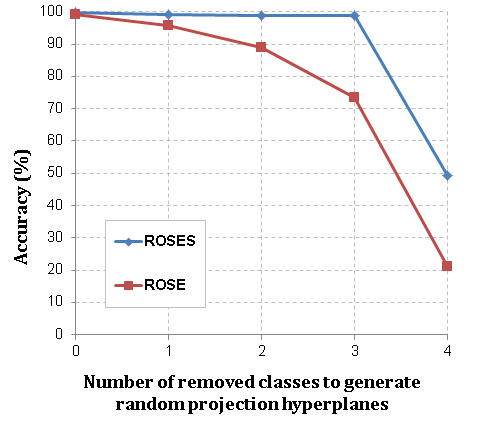}
  \includegraphics[width=1.01\columnwidth,keepaspectratio]{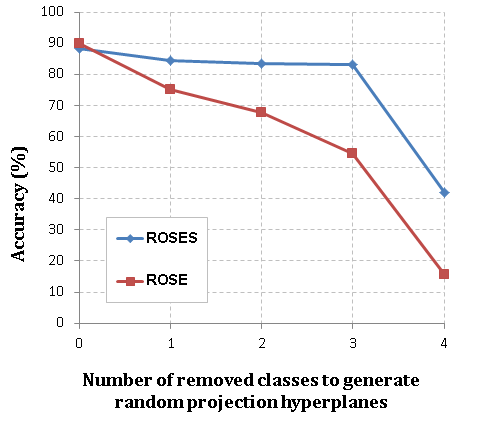}
  \caption
    {
    \small
    Sensitivity of random projection space discriminability to the number of selected data points for generating the random hyperplanes,
    as well as the effect of adding synthetic data points for improving space discriminability.
    The graphs compare the performance of ROSE and ROSES on $5c$ (top) and $5m$ (bottom) sets of the BRODATZ texture recognition dataset respectively.
    }
  \label{fig:ETHZ_SyntheticHighlight}
\end{figure}

To further highlight the proposed ROSES method we set an experiment on two sequences `5c' and `5m' of the \mbox{BRODATZ} dataset. 
In this experiment we reduce the number of data required for creating the mapping function step by step. 
First step we use all the provided training data to construct the random projection space.
Then, we progressively discard training data points from a particular class to construct the space.
We repeat this process until there is only one class left.
Both '5c' and '5m' have a total of 5 classes where each class has 5 samples for training. 
We ran the experiment on every single combination for each case (\eg ~when two classes are excluded, there are 10 combinations) and present the average accuracy.

As shown in Fig.~\ref{fig:ETHZ_SyntheticHighlight},
there is a significant performance difference between the ROSE and ROSES methods,
highlighting the importance of the training data generating the random projection hyperplanes.
This performance difference is more pronounced when more classes are excluded from the training data.
We note that this training set is different from the training set to train the classifier.
Although we exclude some classes in the training set for constructing the random projection space, we still use all the provided training data to train the classifier.

\subsection{Comparison with Recent Methods}
\label{subsec:Experiment_soa}

Table~\ref{tab:FERET_Art} shows that on the FERET face recognition dataset the proposed ROSE method obtains considerably better results
than several recent methods:
log-Euclidean sparse representation (logE-SR)~\cite{SR_Riemannian_AVSS_2010,Yuan:ACCV:2010},
Tensor Sparse Coding~(TSC)~\cite{sivalingam2010tensor},
Locality Preserving Projection~(RLPP)~\cite{harandikernel},
and Relational Divergence Classification~(RDC)~\cite{alavi2013relational}.

\begin{table}[!b]
  \centering
  \caption
    {
    \small
    Recognition accuracy (in \%) for the face recognition task
    using
    log-Euclidean sparse representation (logE-SR)~\cite{SR_Riemannian_AVSS_2010,Yuan:ACCV:2010},
    Tensor Sparse Coding (TSC)~\cite{sivalingam2010tensor},
    Riemannian Locality Preserving Projection~(RLPP)~\cite{harandikernel},
    Relational Divergence Classification (RDC)~\cite{alavi2013relational},
    and the proposed ROSE method.
    }
    \label{tab:FERET_Art}
  \vspace{0.5ex}
 \small
    \begin{tabular}{lccccc}
    \toprule
    &~{\bf LogE-SR}~
    &{\bf TSC}~
    &{\bf RLPP}~
    &{\bf RDC}~
    &{\bf ROSE}~\\
    \toprule
    {\bf bd}     &$35.0$    &$36.0$  &$47.0$ &$59.0$ &${\bf 70.0}$\\
    {\bf bg}   &$47.0$    &$45.0$ &$58.0$ &$71.0$ &${\bf 80.5}$\\
    \bottomrule
        \toprule
{\bf Average}			&$41.0$	&$40.5$	&$52.5$	&$65.0$	&$\bf75.2$\\
    \bottomrule
   \end{tabular}

\end{table}

Table~\ref{tab:BRODATZ_Art} contrasts the performance of the ROSES method (ROSE augmented with synthetic data) on the BRODATZ texture recognition task against the above methods.
We note that in this case the the use of synthetic data is necessary in order to achieve improved performance.
On average, ROSES achieves higher performance than the other methods, with top performance obtained in 7 out of 9 tests.

\begin{table}[!t]
  \centering
  \caption
    {
    \small
    Performance on the Brodatz texture dataset~\cite{randen1999filtering} for
    LogE-SR~\cite{SR_Riemannian_AVSS_2010,Yuan:ACCV:2010}, 
    Tensor Sparse Coding (TSC)~\cite{sivalingam2010tensor},
    Riemannian Locality Preserving Projection~(RLPP)~\cite{harandikernel},
    Relational Divergence Classification (RDC)~\cite{alavi2013relational},
    and the proposed ROSES method.  
    }
    \label{tab:BRODATZ_Art}
    \vspace{0.5ex}
 \small
    \begin{tabular}{lccccc}
    \toprule
    &~{\bf LogE-SR}~
	 &{\bf TSC}~
    &{\bf RLP}~
    &{\bf RDC}~
    &{\bf ROSES}~\\
    \toprule

{\bf 5c}         	    &$89.0$	&$99.7$	&$99.2$	&$98.2$	&$\bf 99.8$ \\
    {\bf 5m}         	&$53.5$	&$72.5$	&$86.2$	&$88.0$	&$\bf 88.4$\\
    {\bf 5v}         	&$73.5$	&$86.3$	&$86.4$	&$87.0$	&$\bf 88.6$\\
    {\bf 5v2}         	&$70.8$	&$86.1$	&$90.0$	&$89.0$	&$\bf 92.7$\\
    {\bf 5v3}         	&$63.6$	&$83.1$	&$89.7$	&$87.0$	&$\bf 91.3$\\
    {\bf 10}         	&$60.6$	&$81.3$	&$84.7$	&$84.0$	&$\bf 87.0$\\
    {\bf 10v}         	&$63.4$	&$67.9$	&$83.0$	&$86.0$	&$\bf 88.5$\\
    {\bf 16c}         	&$67.1$	&$75.1$	&$82.0$	&$\bf 88.0$	&$85.7$\\
    {\bf 16v}         	&$55.4$	&$66.6$	&$74.0$	&$\bf 81.0$	&$79.8$\\
     \bottomrule
    \toprule
{\bf Average}			&$66.3$	&$79.8$	&$86.1$	&$87.6$	&$\bf 89.1$\\
\bottomrule
\end{tabular}

\end{table}

\begin{table}[!t]
  \centering
  \caption
    {
    \small
    Recognition accuracy (in \%) for the person re-identification task on Seq.1 and Seq.2 of the ETHZ dataset. HPE: Histogram Plus Epitome~\cite{HPE_ICPR2010}; SDALF: Symmetry-Driven Accumulation of Local Features~\cite{SDALF_CVPR2010}; RLPP: Riemannian Locality Preserving Projection~\cite{harandikernel}; RDC: Relational Divergence Classification~\cite{alavi2013relational}.
    }
    \label{tab:ETHZ_stateoftheart}
  \vspace{0.5ex}
 \small
    \begin{tabular}{lccccc}
    \toprule
    &{\bf HPE}~
	 &{\bf SDALF}~
    &{\bf RLPP}~
    &{\bf RDC}~
    &{\bf ROSES}~\\
    \toprule
    {\bf Seq.1}       &$79.5$        &$84.1$          &$88.2$     &$88.7$ &${\bf 92.5}$\\
    {\bf Seq.2}        &$85.0$        &$84.0$          &$89.8$     &$89.8$ &${\bf 94.0}$\\
    \bottomrule
        \toprule
{\bf Average}			&$82.2$	&$84.0$	&$89.0$	&$89.2$ &$\bf93.2$\\
    \bottomrule
   \end{tabular}

\end{table}

Finally, we compared the ROSES method with several state-of-the-art algorithms for person re-identification on the ETHZ dataset:
Histogram Plus Epitome~(HPE)~\cite{HPE_ICPR2010},
Symmetry-Driven Accumulation of Local Features~(SDALF)~\cite{SDALF_CVPR2010},
RLPP~\cite{harandikernel}
and RDC~\cite{alavi2013relational}.
The performance of TSC~\cite{sivalingam2010tensor} was not evaluated
due to the method's high computational demands: it would take approximately 200 hours to process the ETHZ dataset.
We do not report the results for LogE-SR due to its low performance on the other two datasets.
The results shown in Table~\ref{tab:ETHZ_stateoftheart} indicate that the proposed ROSES method obtains better performance.
As in the previous experiment, the use of synthetic data is necessary to obtain improved performance.
\section{Main Findings and Future Directions}
\label{sec:conclusions}

The key advantage of representing images in forms of non-singular covariance matrix groups is that superior performance can be achieved when the underlying structure of the group is considered.
It has been shown that when endowed with the Affine Invariant Riemannian Metric (AIRM), the matrices form a connected, smooth and differentiable Riemannian manifold.
Working directly on the manifold space via AIRM poses many computational challenges.
Typical ways of addressing this issue include embedding the manifolds to tangent spaces,
and embedding into Reproducing Kernel Hilbert Spaces (RKHS).
Embedding the manifolds to tangent spaces considerably simplifies further analysis, at the cost of disregarding some of the manifold structure.
Embedding via RKHS can better preserve the manifold structure, but adds the burden of extending existing Euclidean-based learning algorithms into RKHS.

In this work, we have presented a novel solution which embeds the data points into a random projection space by first generating random hyperplanes in RKHS and then projecting the data in RKHS into the random projection space.
We presented a study of space discriminability for various computer vision classification tasks and found that the space 
has superior discriminative power to the typical approaches outlined above.
In addition, we found that the space discriminative power depends on the completeness of training data generating the random hyperplanes. 
To address this issue, we proposed to augment training data with synthetic data.

Experiments on face recognition, person re-identification and texture classification
show that the proposed method (combined with a linear SVM)
outperforms state-of-the-art approaches such as Tensor Sparse Coding, Histogram Plus Epitome, Riemannian Locality Preserving Projection and Relational Divergence Classification.
To our knowledge this is the first time random projection space has been applied to solve classification tasks in manifold space.
We envision that the proposed method can be used to bring superior discriminative power of manifold spaces to more general vision tasks, such as object tracking.

\section*{Appendix}

Here we provide more details to support Proposition 3.3:
we show that for SPD matrices, {\small $\operatorname{log}({\Mat{X}}^c) = c \times \operatorname{log}(\Mat{X})$}.
For this proof we let $c$ be a discrete number, however we note that it can be extended to continuous $c$ .
\begin{small}
\begin{equation}
	\operatorname{log}({\Mat{X}}^c) = \operatorname{log}(\underset{c~\mbox{instances of}~\Mat{X}}{\underbrace{\Mat{X} \times \Mat{X} \times ... \times \Mat{X}}})
\end{equation}%
\end{small}%
Replacing {\small $\Mat{X} \in {Sym}_{+}^{d}$} with its singular value decomposition (SVD) as {\small $\Mat{X} = \Mat{U} \Mat{V} \Mat{U}^{\top}$},
the above equation becomes:
\begin{small}
\begin{equation}
	  \operatorname{log}({\Mat{X}}^c) = \operatorname{log}({\Mat{U} \Mat{V} \Mat{U}^{\top} \times ... \times \Mat{U} \Mat{V} \Mat{U}^{\top}})
\end{equation}%
\end{small}%
As {\small $\Mat{X} \in {Sym}_{+}^{d} $}, the eigenvalue matrix {\small $\Mat{U}$} is orthonormal, and hence {\small $\Mat{U}^{\top}\Mat{U} \mbox{~=~} \Mat{I}$}.
As such, the following equation is valid: 
\vspace{-1ex}
\begin{small}
\begin{equation}
	\operatorname{log}({\Mat{X}}^c) = \operatorname{log}({\Mat{U} \Mat{V}^c \Mat{U}^{\top}})
\end{equation}%
\end{small}%

\noindent
Similarly, as {\small $\Mat{X} \in {Sym}_{+}^{d} $}, {\small $\operatorname{log}(\Mat{X}^c) \mbox{~=~} \Mat{U}\operatorname{log}(\Mat{V}^c)\Mat{U}^{\top}$},
where {\small $\operatorname{log}(\Mat{V}^c)$} is the diagonal matrix of the eigenvalue logarithm~\cite{Tuzel_2008_PAMI}.
Hence we have:
\begin{small}
\begin{eqnarray}
	\operatorname{log}({\Mat{X}}^c)& = &\Mat{U}\operatorname{log}(\Mat{V}^c)\Mat{U}^{\top} \nonumber \\ 
	& = & c\Mat{U}\operatorname{log}(\Mat{V})\Mat{U}^{\top} \nonumber \\ 
	& = & c \times \operatorname{log}(\Mat{X})\nonumber
\end{eqnarray}%
\end{small}%
\section*{Acknowledgements}

This research was partly funded by Sullivan Nicolaides Pathology (Australia),
and the Australian Research Council Linkage Projects Grant LP130100230. 

NICTA is funded by the Australian Government through the Department of Communications and the Australian Research Council through the ICT Centre of Excellence Program.

\renewcommand{\baselinestretch}{1.025}\small\normalsize

\newpage
\footnotesize
\bibliographystyle{ieee}
\bibliography{references}

\end{document}